\documentclass[10pt,twocolumn, letterpaper]{article}

\usepackage{x}
\usepackage{times}
\usepackage{epsfig}
\usepackage{graphicx}
\usepackage{amsmath}
\usepackage{amssymb}
\usepackage[noend]{algpseudocode}
\usepackage{algorithmicx,algorithm}
\usepackage{enumitem}
\usepackage{comment}
\usepackage[table]{xcolor} 
\definecolor{best1}{RGB}{222,242,212}
\definecolor{best2}{RGB}{255,250,212}

\makeatletter
\renewcommand{\paragraph}{%
  \@startsection{paragraph}{4}%
  {\z@}{1ex \@plus 1ex \@minus .2ex}{-1em}%
  {\normalfont\normalsize\bfseries}%
}
\makeatother

\newcommand{\ournerf}{Clean-NeRF\xspace}

\usepackage[pagebackref=true,breaklinks=true,letterpaper=true,colorlinks,bookmarks=false]{hyperref}

\iccvfinalcopy 

\usepackage{xcolor}
\newcommand{\TODO}[1]{\textcolor{red}{#1}}

\begin{document}

\title{Clean-NeRF: Reformulating NeRF to account for View-Dependent Observations}

\author{Xinhang Liu\\
HKUST\\
\and
Yu-Wing Tai\\
HKUST\\
\and
Chi-Keung Tang\\
HKUST
}

\maketitle

\begin{abstract}
While Neural Radiance Fields (NeRFs) had achieved unprecedented novel view synthesis results, 
they have been struggling in dealing with large-scale cluttered scenes with sparse input views and highly view-dependent appearances.
Specifically, existing NeRF-based models tend to produce blurry rendering with the volumetric reconstruction often inaccurate, where a lot of reconstruction errors are observed in the form of foggy ``floaters" hovering within the entire volume of an opaque 3D scene. 
Such inaccuracies impede NeRF's potential for accurate 3D NeRF registration, object detection, segmentation, etc., which possibly accounts for only limited significant research effort so far to directly address these important 3D fundamental computer vision problems to date. 
This paper analyzes the NeRF's struggles in such settings and proposes {\bf \ournerf} for accurate 3D reconstruction and novel view rendering in complex scenes. 
Our key insights consist of enforcing effective appearance and geometry constraints, which are absent in the conventional NeRF reconstruction,  by 1) automatically detecting and modeling view-dependent appearances in the training views to prevent them from interfering with density estimation, which is complete with 2) a geometric correction procedure performed on each traced ray during inference.
\ournerf can be implemented as a plug-in that can immediately benefit existing NeRF-based methods without additional input.
Codes will be released.\footnote{Project page: \hyperlink{ https://xinhangliu.com/cleannerf}{ https://xinhangliu.com/cleannerf}.} 

\end{abstract}

\section{Introduction}
\label{sec:intro}
\begin{figure}[t]
\begin{center}
    \includegraphics[width=1\linewidth]{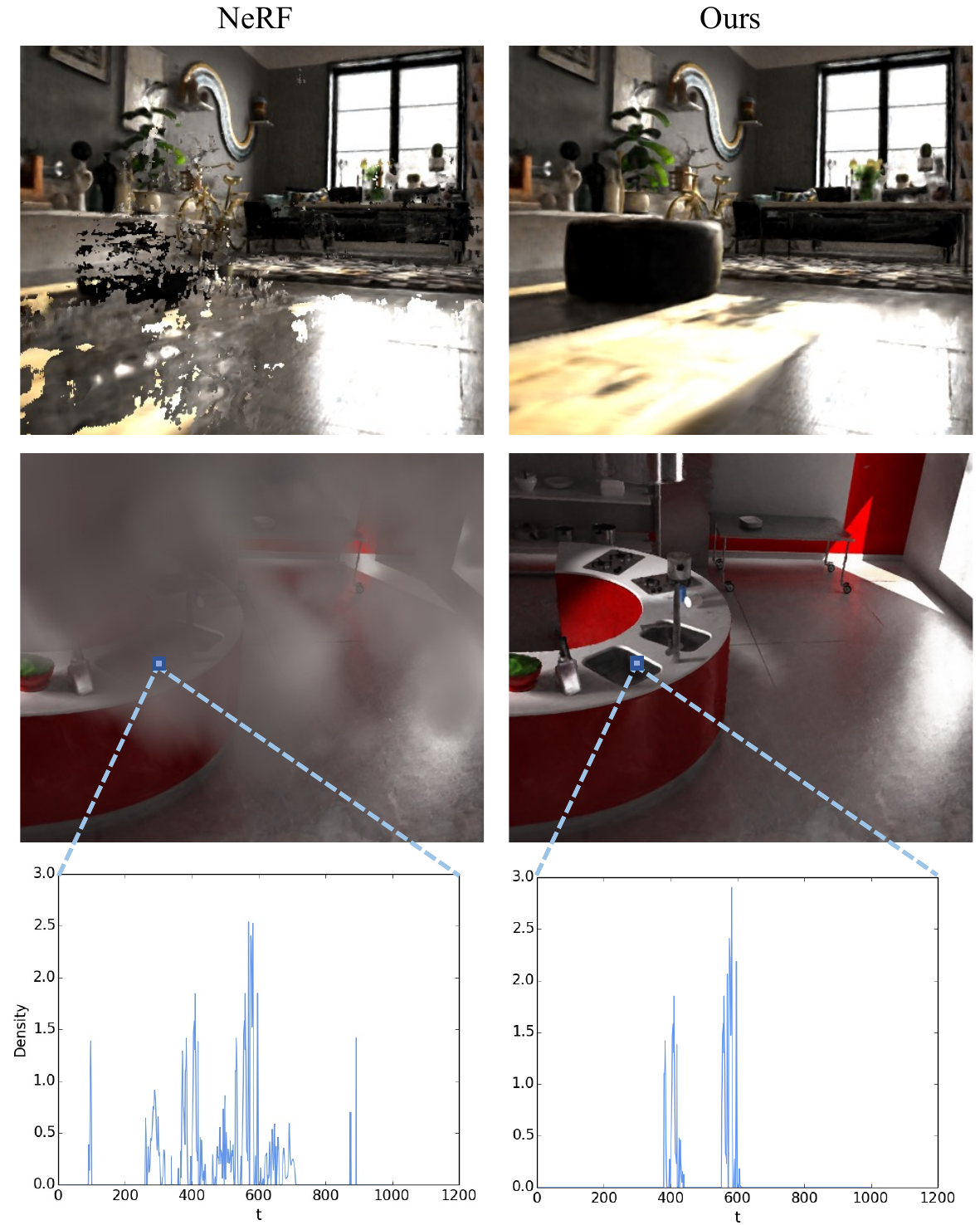}
\end{center}
\vspace{-0.1in}
\caption{\textbf{{\em Foggy} vs {\em Clear} NeRF.} Our \ournerf gets rid of reconstruction errors manifested as foggy ``floaters" in the density volume without additional input or significant computational overhead. 
Below are density profiles along a given ray before and after our geometry correction procedure, where we discard density peaks corresponding to floaters.
}
\label{fig:teaser}
\vspace{-0.2in}
\end{figure}

Neural Radiance Fields (NeRFs)~\cite{mildenhall2020nerf}  
have made revolutionary contributions in 
novel view synthesis~\cite{barron2021mip,barron2022mip}, 
autonomous driving~\cite{rematas2022urban,tancik2022block}, digital human~\cite{hong2022headnerf,zhao2022humannerf}, and 3D content generation~\cite{eg3d,poole2022dreamfusion,lin2022magic3d}.
%
To date, unfortunately, most NeRF-based methods encounter challenges when tackling large-scale cluttered scenes (e.g., Fig.~\ref{fig:teaser}):
\begin{enumerate}[leftmargin=0.16in, topsep=2pt,itemsep=-1ex,partopsep=1ex,parsep=1ex]
\item Input observations used for NeRF are often too sparse  compared to forward-facing or synthetic looking-inward scenes;
\item View-dependent visual effects give rise to ambiguity, resulting in a ``foggy" density field as shown in Fig.~\ref{fig:teaser}. 
Such artifacts are particularly pronounced in indoor scenes strewn with view-dependent appearances, such as specular highlights, glossy surface reflections from man-made objects. 
\end{enumerate}

Despite attempts to enhance NeRF's rendering quality given suboptimal input, such as using 3D conical frustums~\cite{barron2021mip,barron2022mip}, physically-grounded augmentations~\cite{chen2022aug}, and misalignment correction~\cite{jiang2022alignerf},  these challenges have yet to be fully resolved.
Depth supervision~\cite{deng2022depth, wei2021nerfingmvs} or proxy geometry~\cite{xu2021scalable,wu2022scalable} images can help alleviate the challenges in handling large-scale with sparse input, at the expense of 
expensive pre-processing or additional input.
Another line of work~\cite{wang2021neus, oechsle2021unisurf, wang2022neuris} achieves better reconstruction of surface geometry by using signed distances instead of volume density as scene representation. However, they sacrifice the ability to synthesize photo-realistic novel views.

%
%
To address the above issues, we propose an extension to NeRF, dubbed as {\bf \ournerf}, which enforces effective {\em appearance} and {\em geometry} constraints conducive to accurate colors and 3D densities estimation. We believe \ournerf can contribute beyond novel view synthesis, such as NeRF object detection~\cite{hu2022nerf}, NeRF object segmentation~\cite{zhi2021place, liu2022unsupervised, fan2022nerf,ren2022neural}, and NeRF registration~\cite{goli2022nerf2nerf}, where the rooms for improvement are substantial if more accurate color and density estimation are available.

\begin{figure}[t]
\begin{center}
    \includegraphics[width=1\linewidth]{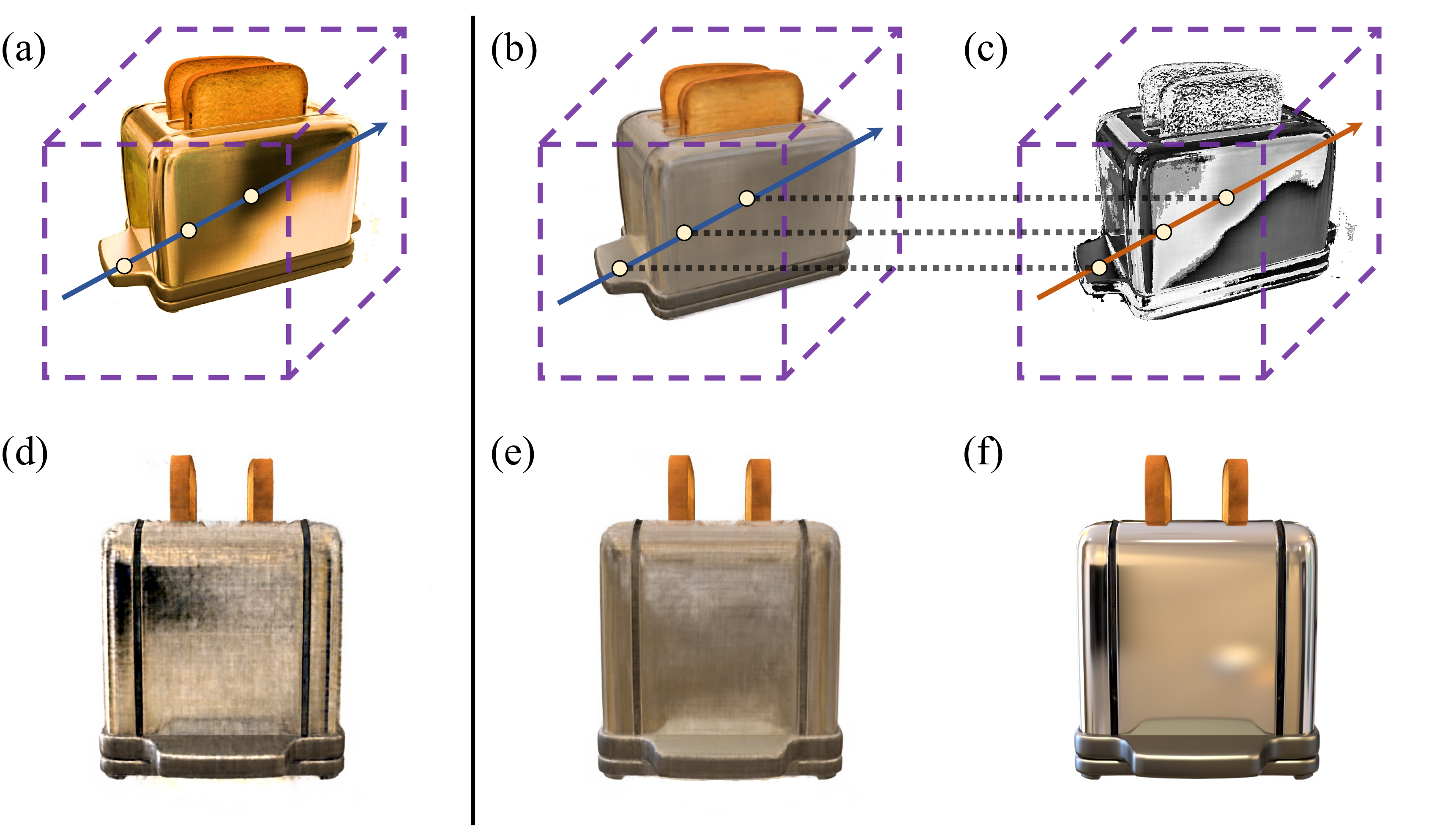}
\end{center}
\vspace{-0.1in}
\caption{\textbf{Appearance Decomposition.} \textbf{(a, d)} For object surfaces with strong view-dependent effects, vanilla NeRF often gets disrupted, resulting in poor reconstruction quality. 
We propose decomposing appearance into \textbf{(b)} view-independent and \textbf{(c)} view-dependent components during NeRF training and incorporating geometry-related priors to improve reconstruction quality. Our approach shows the ability to \textbf{(e)} render the view-independent component of the scene and \textbf{(f)} more accurate recovery of the scene. 
}
\label{fig:toaster}
\vspace{-0.15in}
\end{figure}

Correspondingly, there are two steps in \ournerf. First, for appearance correction, the view-independent and view-dependent color components are predicted from the underlying 3D scene, which is combined to produce the final color estimation (Fig.~\ref{fig:toaster}).
The view-independent component (diffuse color and shading) captures the overall scene color, while the view-dependent component (highlights or reflections) captures color variations due to changes in viewing angle.
\ournerf then discards these view-dependent appearances in the training views to prevent them from interfering with the density estimation.
Second, a simple and effective geometry correction procedure will be performed to further eliminate the foggy ``floaters" or density errors. This geometry correction procedure is based on an assumption in line with traditional ray tracing in computer graphics.
%
Experiments verify that our proposed \ournerf can effectively get rid of floater artifacts without additional input.

In summary, our contributions include the following:
\begin{itemize}[leftmargin=0.16in, topsep=2pt,itemsep=-1ex,partopsep=1ex,parsep=1ex]
    \item We propose a concise method for decomposing view-independent and view-dependent appearance during NeRF training and eliminate the interference of view-dependent appearance.
    \item We propose a geometric correction procedure performed on each traced ray during inference to refine the density estimation and better tackle the floater artifacts.
    \item Extensive experiments and ablations verify the effectiveness of our core designs and results in improvements over the vanilla NeRF and other state-of-the-art alternatives.
\end{itemize}

\section{Related Works}

\paragraph{Neural Radiance Fields.} 
NeRFs debut in~\cite{mildenhall2020nerf}  achieved unprecedented novel view synthesis effects, by modeling the underlying 3D scene as a continuous volumetric field of color and density via layers of MLP. The input to a NeRF consists of a 5D vector containing a 3D location $(x, y, z)$, and a 2D viewing direction $(\theta, \phi)$. 
Since then, follow-up works have addressed the limitations and improved the performance, such as 
enabling dynamic modeling~\cite{zhang2021stnerf, park2021nerfies, dnerf, tretschk2021non},
efficient inference~\cite{yu2021plenoctrees, yu2021plenoxels, chen2022tensorf, lombardi2021mixture, muller2022instant, fpo_cvpr}, 
editing~\cite{liu2021editing, zhang2021stnerf, wang2021clip, jang2021codenerf,kobayashi2022decomposing}, and
multi-task learning~\cite{zhi2021place,liu2022unsupervised,fan2022nerf}.

Much  effort has been made to improve the accuracy of NeRF's rendered novel views and reconstructed geometry.
Mip-NeRF~\cite{barron2021mip} mitigates unsightly aliasing artifacts in NeRF in representing coarse and fine details.
Mip-NeRF 360~\cite{barron2022mip} uses a non-linear scene parameterization and online distillation to tackle unbounded scenes.
Aug-NeRF~\cite{chen2022aug} leverages worst-case perturbations to enable NeRF in smoothness-aware geometry reconstruction.
RegNeRF~\cite{Niemeyer2021Regnerf}  renders patches from unobserved viewpoints for a given radiance field and regularizes appearance and geometry.
AligNeRF~\cite{jiang2022alignerf}  addresses misalignment problems caused by moving objects or small camera calibration errors.
DS-NeRF~\cite{deng2022depth}  takes advantage of readily-available depth supervision from SfM.
NerfingMVS~\cite{wei2021nerfingmvs} employs  adapted depth priors from a monocular depth network to guide the sampling process of volume rendering.

\paragraph{Reflectance Decomposition.} To acquire reflectance data, sophisticated devices have traditionally been necessary to sample the light-view space~\cite{kang2018efficient, matusik2003data, nielsen2015optimal}.
Subsequent research has proposed practical techniques for acquiring spatially varying BRDFs, such as those presented in~\cite{kang2018efficient, matusik2003data, nielsen2015optimal, nam2018practical}.
More recently, deep learning methods have made it possible to acquire BRDF information from a single flash image~\cite{li2018learning,li2018materials,deschaintre2018single}.

In the context of NeRF, highly reflective objects can pose challenges in the reconstruction and relighting process. Previous works have attempted to address this issue by decomposing appearance into scene lighting and materials, but these methods assume known lighting~\cite{bi2020b, srinivasan2021nerv} or no self-occlusion~\cite{boss2021nerd, boss2021neural, zhang2021physg}.
Ref-NeRF~\cite{verbin2022ref} uses a representation of reflected radiance and structures this function using a collection of spatially-varying scene properties to reproduce the appearance of glossy surfaces.
Despite these advances, Ref-NeRF requires accurate normal vectors and outgoing radiance estimation, which is difficult to obtain for sparse input views.
In addition, effectively addressing the view-dependent appearance problem in the context of large scenes and sparse observations remains a challenge. To address this issue, we propose a simple yet effective decomposition method to eliminate its interference without the need to estimate surface normals or outgoing radiance.

\paragraph{Intrinsic Image Decomposition.}
Barrow and Tenenbaum introduced intrinsic images as a valuable intermediate representation for scenes~\cite{barrow1978recovering}, assuming that an image can be expressed as the pointwise product of the object's true colors or reflectance and the shading on that object. This can be represented as
$
I = R \cdot S,
$
where $I$, $R$, and $S$ denote the image, the reflectance, and the shading, respectively.

Early optimization-based works addressed the problem of separating an image into its reflectance and illumination components by assuming that large image gradients correspond to reflectance changes and small gradients to lighting changes~\cite{land1971lightness, horn1974determining}.
Incorporation of additional priors improves the accuracy and robustness, such as 
reflectance sparsity~\cite{Rother2011RecoveringIntrinsic, Shen2011IntrinsicImages}, 
low-rank reflectance~\cite{Bousseau2015User_assisted} and  distribution difference in gradient domain~\cite{Bi2015IntrinsicDecompositionTOG, Li2014Single}.
Deep learning methods~\cite{fan2018revisiting,yu2019inverserendernet, zhu2022irisformer, li2018cgintrinsics, li2018learning_intrinsic} have emerged to perform intrinsic image decomposition, estimating the reflectance and shading on labeled training data.
Notably and differently, in intrinsic image decomposition, where shadows and highlights are separated as high-frequency components, these components {\em may} still be separated as view-independent in our \ournerf as long as they are {\em consistent} across all input views, e.g., a static shadow is consistently observed across all views. Thus, intrinsic image decomposition is inappropriate (both overkill and inadequate) 
to the ``vi-vd" decomposition of \ournerf.

IntrinsicNeRF~\cite{ye2022intrinsicnerf} introduces intrinsic decomposition to the NeRF-based neural rendering method, which allows for editable novel view synthesis in room-scale scenes. Compared with our simple and effective appearance decomposition, IntrinsicNeRF requires dense inputs (900 images for their indoor Replica scene), which  assumes the NeRF reconstruction is accurate.

\section{Method}
In this section, we present \ournerf to effectively 
address the floater artifacts caused by current inaccurate NeRF reconstruction.
%
\ournerf can be applied as an easy plug-in component 
into any existing NeRF-like models, including MLP-based and voxel grid-based approaches, without resorting to  extra knowledge such as normal maps, depth maps, or lighting conditions.
In Sec.~\ref{sec:overall}, we introduce the overall architecture of \ournerf compared to the vanilla NeRF.
Sec.~\ref{sec:vivd} describes our principled implementation for decomposing an appearance into view-dependent (``vd") and view-independent (``vi") components to achieve better novel view synthesis and 3D geometry reconstruction.
Finally, we present our geometry correction strategy  in Sec.~\ref{sec:geo}. 

\begin{figure}[t]
\begin{center}
    \includegraphics[width=1\linewidth]{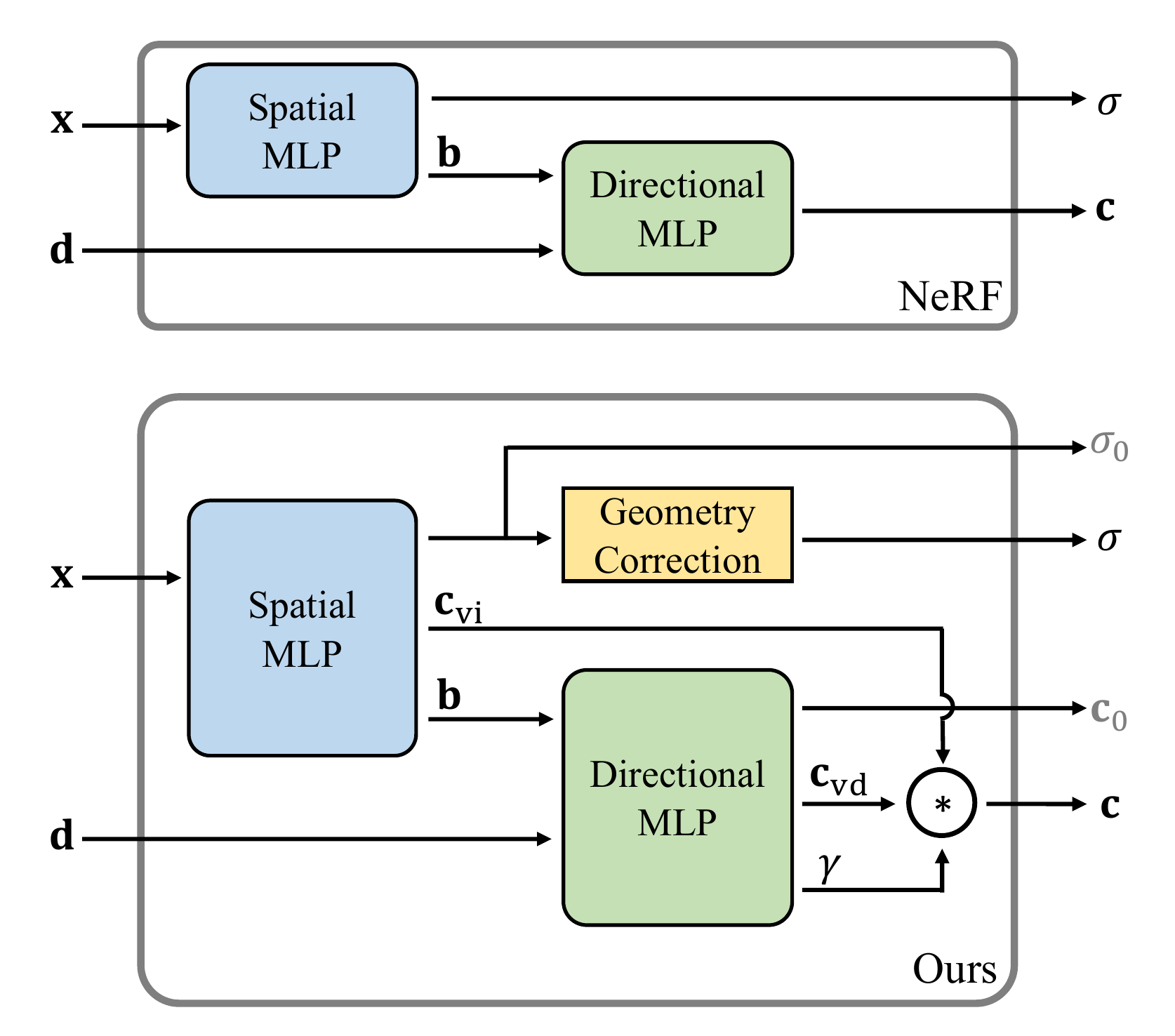}
\end{center}
\vspace{-0.1in}
\caption{\textbf{Vanilla NeRF and \ournerf architectures. }
Similar to vanilla NeRF, we estimate density $\sigma_0$ and color $\mathbf{c}_0$ from the spatial MLP and the directional MLP.
On the other hand, we propose ``vi-vd" appearance decomposition based on spherical harmonics (SHs) and obtain a final color estimation $\mathbf{c}$ by summing the view-independent color component $\mathbf{c}_{\text{vi}}$ produced by the spatial MLP, with the view-dependent color component $\mathbf{c}_{\text{vd}}$ produced by the directional MLP.
Then a geometry correction procedure refines the initial estimation and obtains the final density estimation $\sigma$.
}
\label{fig:method}
\vspace{-0.15in}
\end{figure}

\subsection{Overall Architecture}
\label{sec:overall}
Fig.~\ref{fig:method} compares  vanilla NeRF~\cite{mildenhall2020nerf} and our model, both taking a sampled spatial coordinate point $\mathbf{x} = (x, y, z)$ and direction $\mathbf{d} = (\theta, \phi)$ as input and output the volume density and color.

Vanilla NeRF uses a spatial MLP to estimate volume density $\sigma$ at position $\mathbf{x}$.
Then the directional MLP takes as input the direction $\mathbf{d}$ as well as spatial feature $\mathbf{b}$ to estimate a view-depenent output color $\mathbf{c}$.
In \ournerf architecture, we also use the spatial MLP and the directional MLP to output an initial estimation of density $\sigma_0$ and color $\mathbf{c}_0$, similar to the vanilla NeRF. 
The initial estimation of color corresponding to ray $\mathbf{r}(t)=\mathbf{o}+t\mathbf{d}$ can be evaluated from  $\sigma_0$ and $\mathbf{c}_0$:
\begin{equation}
\label{eqn:vanillanerf}
    \hat{\mathbf{C}}_0 = \sum_{k=1}^K \hat{T}_0(t_k)\alpha(\sigma_0(t_k)\delta_k)\mathbf{c}_0(t_k), 
\end{equation}
where $\hat{T}_0(t_k)=\operatorname{exp}\left(-\sum_{k^{'}=1}^{k-1}\sigma_0(t_k)\delta(t_k)\right)$, $\alpha \left({x}\right) = 1-\exp(-x)$, and $\delta_p = t_{k+1} - t_k$.
Without requiring additional inputs, \ournerf differs from vanilla NeRF in that while making the initial estimation, we also predict the view-independent color component $\mathbf{c}_{\text{vi}}$ with the spatial MLP and the view-dependent color component $\mathbf{c}_{\text{vd}}$  with the directional MLP.
We supervise these estimations by performing an SH-based decomposition of the initial color estimation $\mathbf{c}_0$, as described in Sec.~\ref{sec:vivd}.
The directional MLP  also estimates a view-dependent factor $\gamma$ and we obtain a final color estimation $\mathbf{c}$ by:
\begin{equation}
\label{eqn:blend}
    \mathbf{c} = \gamma\mathbf{c}_{\text{vi}} + (1-\gamma)\mathbf{c}_{\text{vd}}.
\end{equation}
Note that $\mathbf{c}_{\text{vi}}$ captures the overall scene color while $\mathbf{c}_{\text{vd}}$ captures the color variations due to changes in viewing angle.
By blending these two components with the factor $\gamma$, we can synthesize a faithful color of the underlying 3D scene, even from a limited number of input views.
To better eliminate the floating artifacts, we propose a geometry correction procedure on $\sigma_0$ to correct the initial density estimation and obtain the final density estimation $\sigma$, as described in Sec.~\ref{sec:geo}.
Such procedure is based on practical assumptions and is easy to implement.

The final estimation of color corresponding to ray $\mathbf{r}(t)=\mathbf{o}+t\mathbf{d}$ is then computed as
\begin{equation}
\label{vol_rend}
    \hat{\mathbf{C}} = \sum_{k=1}^K \hat{T}(t_k)\alpha(\sigma(t_k)\delta_k)\mathbf{c}(t_k), 
\end{equation}
where $\hat{T}(t_k)=\operatorname{exp}\left(-\sum_{k^{'}=1}^{k-1}\sigma(t_k)\delta(t_k)\right)$, $\alpha \left({x}\right) = 1-\exp(-x)$, and $\delta_p = t_{k+1} - t_k$.
We train the network using photometric loss based on both the initial and the final estimations
\begin{equation}
    \mathcal{L}_{\text{pho}} =\sum_{\mathbf{r}\in\mathcal{R} } \left\|\hat{\mathbf{C}}_0 (\mathbf{r}) - \mathbf{C}(\mathbf{r}) \right\|_2^2 + \left\|\hat{\mathbf{C}}(\mathbf{r}) - \mathbf{C}(\mathbf{r}) \right\|_2^2.
\end{equation}

\subsection{Appearance Decomposition}
\label{sec:vivd}
To guide our vi-vd decomposition, we utilize Spherical harmonics (SHs) which are widely used as a low-dimensional representation for spherical functions, and have been used to model Lambertian surfaces \cite{ramamoorthi2001relationship, basri2003lambertian} as well as glossy surfaces \cite{sloan2002precomputed}. 
To use SH functions to model a given function, we query the SH functions $Y_{\ell}^m:\mathbb{S}^2\mapsto\mathbb{R}$ at a viewing angle $\mathbf{d}$ and then fit the estimation $\mathbf{c}_0$ by finding the corresponding coefficients. 
We use low-degree SH functions to compute ideal values of view-independent color components, and high-degree SH functions for view-dependent components.
In this subsection, we will perform all of our calculations at an arbitrary position $\mathbf{x}$ in space, and therefore we will omit the symbol $\mathbf{x}$ from our notation.

We use  $\mathbf{y}(\mathbf{d})\in\mathbb{R}^{L}$  to represent the set of SH function values at the viewing angle $\mathbf{d}$:
\begin{equation}
    \mathbf{y}(\mathbf{d}) = \left[
        Y_{0}^0(\mathbf{d}),Y_{1}^{-1}(\mathbf{d}), Y_{1}^{0}(\mathbf{d}),Y_{1}^{1}(\mathbf{d}), \dots, Y_{\ell_{\text{max}}}^{\ell_{\text{max}}}(\mathbf{d})\right]^\top,
\end{equation}
where $L=(\ell_{\text{max}}+1)^2$. 
To ensure clarity, we will use $c:\mathbb{S}^2\mapsto\mathbb{R}$ to represent one of the three channels of $\mathbf{c}_0$ at a given position $\mathbf{x}$ (also $c_{\text{vi}}$ and $c_{\text{vd}}$), noting the derivation should be readily extended to all three channels. 
We begin by sampling a set of $N$ viewing angles ${\mathbf{d_i}}, {1\leq i\leq N}\subset\mathbb{S}^2$.
The colors of all the sample directions are represented using a vector $\mathbf{s}\in\mathbb{R}^{N}$:
\begin{equation}
    \mathbf{s} =\begin{bmatrix}c(\mathbf{d}_1)&c(\mathbf{d}_2)&\dots& c(\mathbf{d}_{N-1})&c(\mathbf{d}_N)\end{bmatrix}^\top
\end{equation}
The coefficients to be determined are represented by a vector $\mathbf{k}\in\mathbb{R}^{L}$. To find the optimal coefficients that fit the view-dependent color estimation, we solve the following optimization problem:
\begin{equation} 
    \min_{\mathbf{k}\in\mathbb{R}^L} \|\mathbf{s}-\mathbf{Y}\mathbf{k}\|_2^2,
\end{equation}
where 
\begin{equation}
    \mathbf{Y} = \begin{bmatrix}
        \mathbf{y}(\mathbf{d}_1) & \mathbf{y}(\mathbf{d}_2)&\dots&\mathbf{y}(\mathbf{d}_{N-1})&\mathbf{y}(\mathbf{d}_N)
    \end{bmatrix}.
\end{equation}
This is a standard linear regression problem, where we seek to find the values of the coefficient vector $\mathbf{k}$ which minimizes the least squares error between the vector $\mathbf{s}$ and the linear combination of the columns of $\mathbf{Y}$, weighted by the coefficients in $\mathbf{k}$. Using the normal equation, the solution is given by:
\begin{equation}
    \mathbf{k}^* = (\mathbf{Y}^\top\mathbf{Y})^{-1}\mathbf{Y}^\top\mathbf{s}
\end{equation}
We can use the solution coefficients $\mathbf{k}^*$ as weights to linearly combine SH functions. Retaining the low-degree SH functions allows us to capture the view-independent appearance of the scene. Conversely, including high-degree SH functions leads to a high-frequency view-dependent residue. To differentiate between the two, we denote the low-degree and high-degree functions as $L_{\text{low}}$ and $L_{\text{high}}$, respectively.
To compute ideal values for the view-independent component, we can use the solution coefficients and apply the following equation:
\begin{equation}
\tilde{c}_\text{vi}=\sum_{i=1}^{L_\text{low}} \frac{1}{4\pi r^2} \iint_{\mathbb{S}^2} k^*_{i}y_i(\mathbf{d}) \sin\theta d\theta d\phi,
\end{equation}
where we take the mean value around the $\mathbb{S}^2$ surface. 
In our implementation, we approximate it by 
\begin{equation}
\tilde{c}_\text{vi}=\sum_{i=1}^{L_\text{low}} \sum_{i=1}^N k^*_{i}y_i(\mathbf{d_i}),
\end{equation}
This value is then used to guide the output of the view-independent color component $\mathbf{c}_{\text{vi}}$ from the spatial MLP using a regularizer. Specifically, we use the following equation to compute the vi-regularizer loss $\mathcal{L}_{\text{vi}}$:
\begin{equation}
\label{eqn:lvi}
\mathcal{L}_{\text{vi}} = (\mathbf{c}_{\text{vi}}- \tilde{\mathbf{c}}_{\text{vi}})^2.
\end{equation}
We apply the following equation to compute optimal values for the view-dependent component:
\begin{equation}
\tilde{c}_\text{vd}(\mathbf{d})=\sum_{i=L_\text{high}}^{L} k^*_{i}y_i(\mathbf{d}).
\end{equation} 
Incorporating the computed value to guide the output of the view-dependent color residue $\mathbf{c}_{\text{vd}}$ from the directional MLP using the vd-regularizer loss:
\begin{equation}
\label{eqn:lvd}
    \mathcal{L}_{\text{vd}} = \left\|
    \begin{bmatrix}
        \mathbf{c}_{\text{vd}}(\mathbf{d_1}) \\ \vdots \\ \mathbf{c}_{\text{vd}}(\mathbf{d_N})
    \end{bmatrix} - 
    \begin{bmatrix}
        \tilde{\mathbf{c}}_{\text{vd}}(\mathbf{d_1}) \\ \vdots \\\tilde{\mathbf{c}}_{\text{vd}}(\mathbf{d_N})
    \end{bmatrix}\right\|_2^2.
\end{equation}
As aforementioned we consider a given position $\mathbf{x}$ in the space, while in actual implementation we take the $\ell_2$-norm among all the positions in a sampled batch for Eqn. \ref{eqn:lvi} and Eqn. \ref{eqn:lvd}.

\subsection{Geometry Correction}
\label{sec:geo}
\begin{figure}[t]
\vspace{-0.2in}
\begin{center}
    \includegraphics[width=1\linewidth]{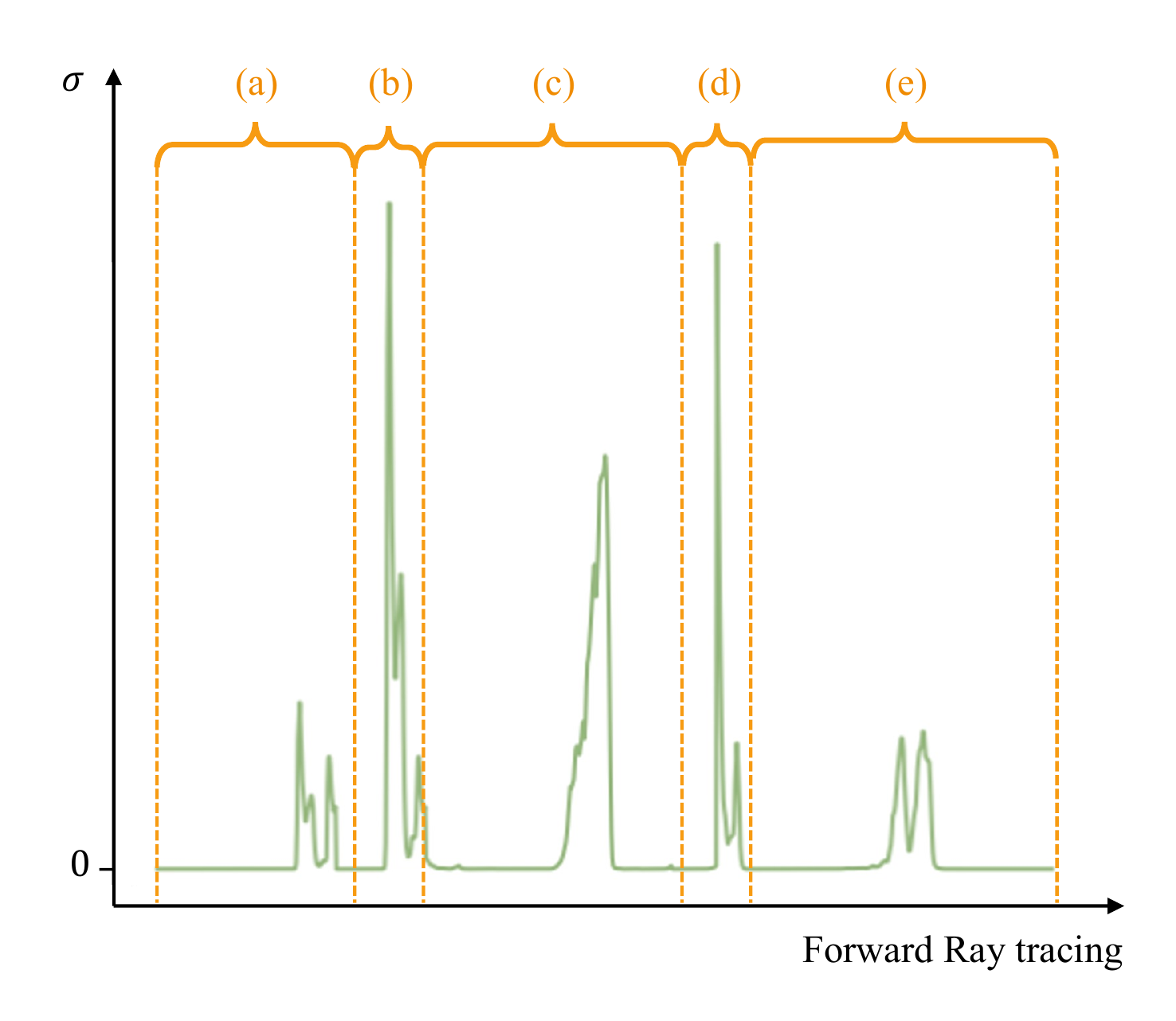}
\end{center}
\vspace{-0.25in}
\caption{\textbf{An example traced ray's density profile. }
Peaks in region (a) and region (e) appear as floating artifacts and would interfere with the rendering process.
}
\label{fig:density}
\vspace{-0.2in}
\end{figure}
\ournerf  is consistent with standard NeRF in that  an initial density estimation, denoted as $\sigma_0$, is generated for volume rendering with the initial color estimation $\mathbf{c}_0$. 
We propose a geometry correction strategy for the final rendering that simultaneously refine the density estimation while better handling  unsightly floater artifacts.
%
%

To understand our working principle, without losing generality, consider a given ray intersecting an opaque 3D scene consisting of a single sphere. In the absence of noise, there exist at most two relevant  intersections, corresponding to the front and back  surface visible from the camera at both ray directions. This is in line with traditional ray tracing in computer graphics, where only the ray-object intersection closest to the camera should be returned.  Figure~\ref{fig:density} shows a density profile along a given ray, where the two salient peaks respectively correspond to the closest front and back surface visible along the ray in both camera view directions.
Other peaks along the same ray 
will either manifest as foggy floaters hovering in the density volume or correspond to surfaces in the scene that are hidden by the front and back surfaces and, thus should not be visible or rendered at all.

Thus, our geometry correction strategy given by Alg.~\ref{alg:geo} begins the density decomposition process by performing a forward ray tracing to identify the first salient peak $k_{\text{peak}}^{\text{front}}$ from the front-facing direction, followed by a backward ray tracing pass to identify the first salient peak $k_{\text{peak}}^{\text{back}}$ from the back-facing direction.
We then retain the density component within and between the neighborhoods of $k_{\text{peak}}^{\text{front}}$ and $k_{\text{peak}}^{\text{back}}$ and zero the rest\footnote{With abuse of language, salient plateaus are also referred to as ``peaks'', which are also detected using Alg.~\ref{alg:geo} if the pertinent ray hits a solid (sphere).}.

\begin{algorithm}[ht]
\caption{Geometry Correction} 
\hspace*{0.02in} {\bf Input:} 
 $\{\sigma_0(t_k)\}_{k=1,...,K}$, $\sigma_{\text{thres}}, m$ \\ 
\hspace*{0.02in} {\bf Output:} 
 $\{\sigma(t_k)\}_{k=1,...,K}$
\begin{algorithmic}[1]

\For{$k\leftarrow1 \text{ to } K$} \hfill // Forward pass ray tracing.
\If{$\sigma_0(t_k)>\sigma_{\text{thres}}$}
\State $k_{\text{peak}}^{\text{front}} \leftarrow k$
\State \textbf{break}
\EndIf
\EndFor

\For{$k\leftarrow K \text{ to } 1$}\hfill // Backward pass ray tracing.
\If{$\sigma_0(t_k)>\sigma_{\text{thres}}$}
\State $k_{\text{peak}}^{\text{back}} \leftarrow k$
\State \textbf{break}
\EndIf
\EndFor

\For{$k\leftarrow 1 \text{ to } K$} 
\If{$k<k_{\text{peak}}^{\text{front}}-m$ \textbf{or} 
$k>k_{\text{peak}}^{\text{back}}+m$  }
\State $\sigma(t_k)\leftarrow 0$
\Else 
\State $\sigma(t_k)\leftarrow \sigma_0(t_k)$
\EndIf
\EndFor
\Return  $\{\sigma(t_k)\}_{k=1,...,K}$
\end{algorithmic}
\label{alg:geo}
\end{algorithm}

\begin{figure*}[t]
\begin{center}
    \includegraphics[width=\linewidth]{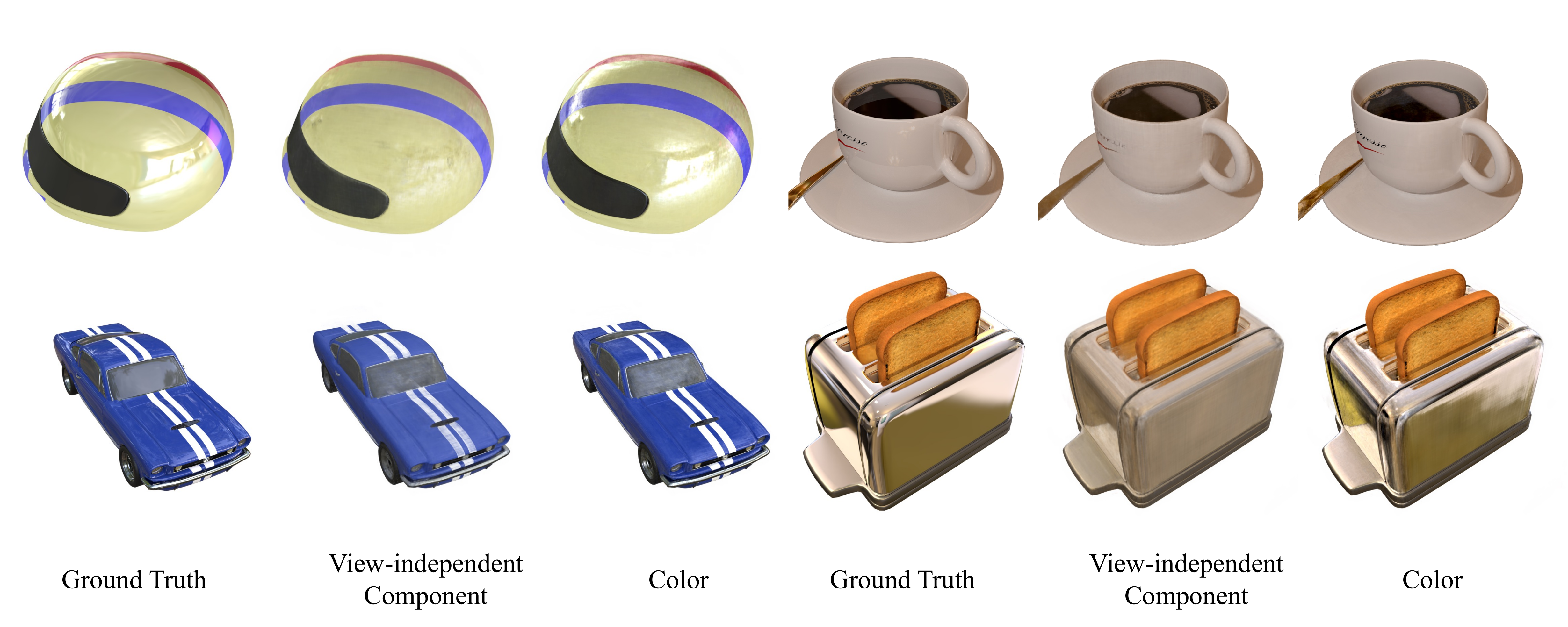}
\end{center}
\vspace{-0.15in}
\caption{\textbf{Qualitative evaluation of \ournerf on Shiny Blender dataset.} We render the view-independent component image, and the final color image combining both view-independent and view-dependent components to compare with the ground truth.
}
\label{fig:results_ref}
\vspace{-0.1in}
\end{figure*}

\begin{figure*}[ht]
\begin{center}
    \includegraphics[width=\linewidth]{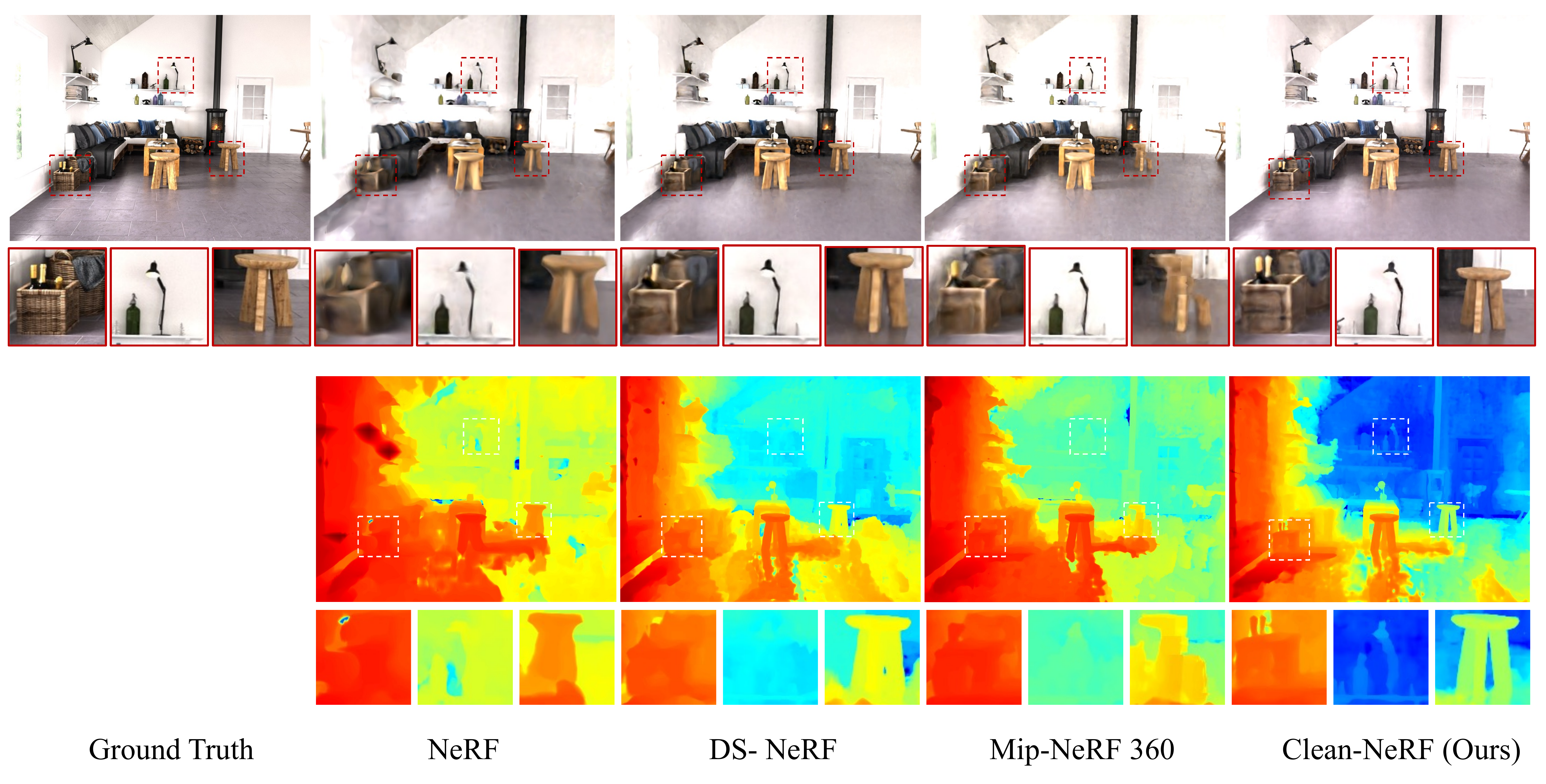}
\end{center}
\vspace{-0.15in}
\caption{\textbf{Qualitative Comparison between \ournerf and baselines.} \ournerf recovers intricate object details while removing annoying ``floaters" even when observations are sparse. 
}
\label{fig:results_indoor}
\vspace{-0.15in}
\end{figure*}

Refer to Fig.~\ref{fig:density} again: peaks in region (a) and region (e) appear as floating artifacts and would interfere with the rendering process, and so they are discarded.
Notably, multiple salient peaks may exist corresponding to other surface points along the ray, such as region (d),  e.g., corresponding to the two slices of toast intersected by the pertinent ray in the previous figure.
If the salient peak (d) is further from the ray origin but lower than (b), as Fig.~\ref{fig:density} shows, the corresponding surface point is occluded by (b), which can be safely detected by backward pass while occlusion-correct rendering is unaffected, as peak (d) is lower. 
%
Otherwise, suppose the peak (d) is higher (e.g., the toast further back is higher), then the peak in (b) should have been clamped to reveal the true geometry (d) as the first salient peak as seen from the ray origin.

\begin{figure*}[t]
\begin{center}
    \includegraphics[width=0.95\linewidth]{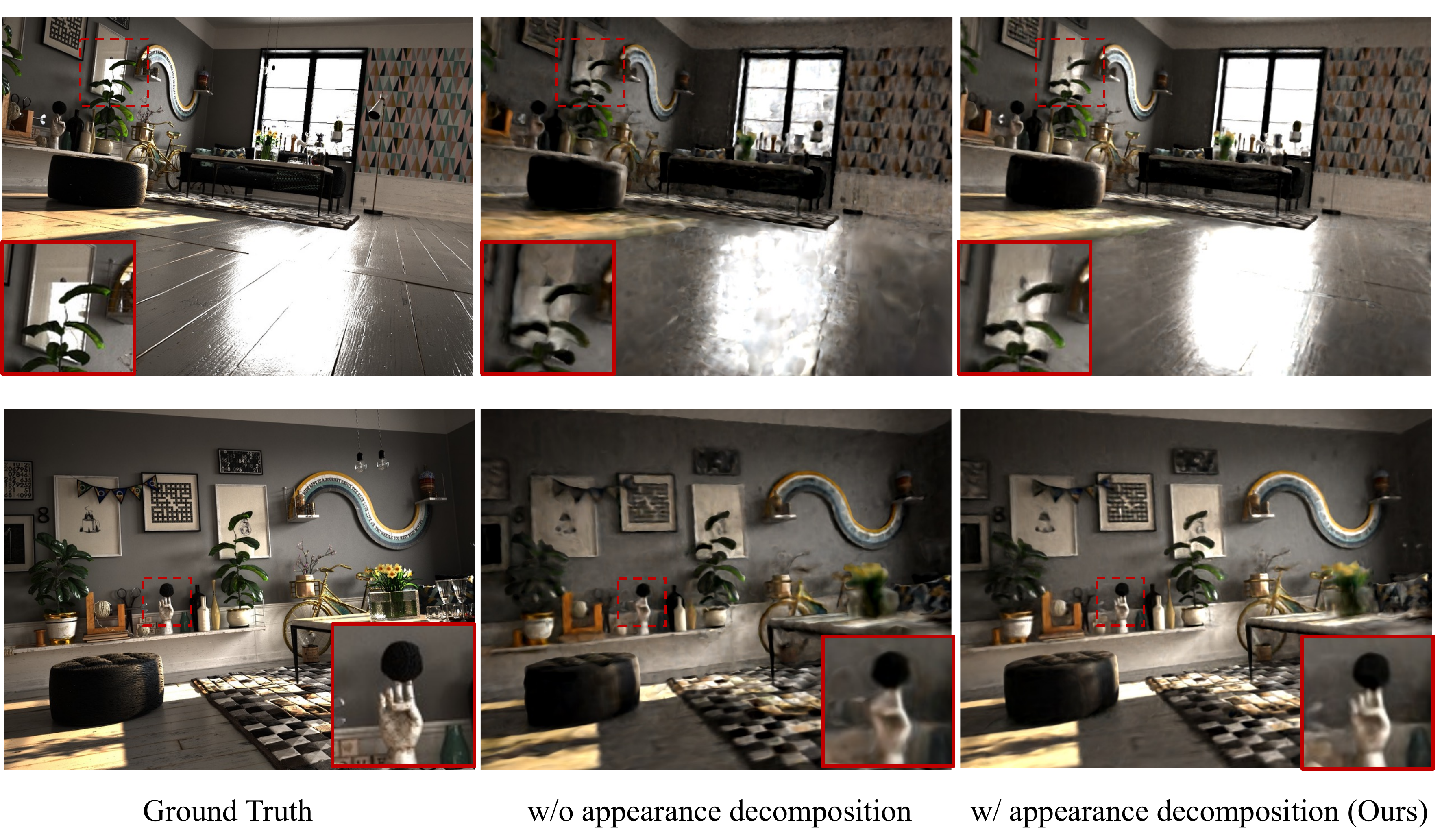}
\end{center}
\vspace{-0.1in}
\caption{\textbf{Qualitative evaluation on appearance decomposition.} Without appearance decomposition, our model fails to recover the glossy objects such as the glass, the floor, and the plant.
}
\label{fig:ab_app}
\vspace{-0.1in}
\end{figure*}

\section{Results}
\label{sec:results}

In this section, we provide comparisons with previous state-of-the-art NeRF-based methods and  evaluation of our main technical components, both qualitatively and quantitatively.
We first evaluate our proposed \ournerf on Shiny Blender dataset~\cite{verbin2022ref} containing different glossy objects rendered in Blender under conditions similar to typical NeRF's datasets, to verify our model's ability to handle challenging material properties by proper decomposition of observed appearance into the corresponding view-independent and view-dependent components.
We then run our proposed \ournerf on Hypersim dataset~\cite{roberts2021hypersim}, a challenging synthetic dataset of photo-realistic indoor scenes.
We use the given accurate camera poses for Shiny Blender and use colmap~\cite{schonberger2016structure} to estimate camera poses for Hypersim.
We train our \ournerf for 500K iterations to guarantee convergence on a single NVIDIA GeForce RTX 3090 Ti GPU. 
For Shiny Blender, we use 100 input views for training. For Hypersim, we use around 50 to 80 input views for training, depending on the scale of the scene. All the shown cases and reported metrics are from held-out views.
We report three error metrics, including peak signal-to-noise ratio (PSNR), structural similarity index measure (SSIM)~\cite{wang2004image}, mean absolute error (MAE), and learned perceptual image patch similarity (LPIPS)~\cite{zhang2018unreasonable}. 
We  include more experiment details and results in the supplemental material, including videos, and encourage the readers to check them.

\paragraph{Appearance Decomposition on Glossy Objects}
To verify \ournerf's ability to decompose object appearances into the corresponding view-independent and view-dependent components, we evaluate \ournerf on Shiny Blender dataset and render the view-independent component image and  color image with both view-independent and view-dependent components (Fig.~\ref{fig:results_ref}).

\paragraph{Comparison on Challenging Indoor Scenes}
We compare \ournerf with NeRF~\cite{mildenhall2020nerf}, mip-NeRF 360~\cite{barron2022mip} and DS-NeRF~\cite{deng2022depth}, which are representative NeRF-based methods and strong baselines for large-scale scenes. As  shown in Fig.~\ref{fig:results_indoor} and Tab~\ref{table:eval2}, our method recovers intricate details of objects in the indoor scene, especially objects with glossy surfaces, and has few observations.
Note that DS-NeRF uses sparse depth maps for supervision, while our method and other baselines do not.

\begin{figure*}[t]
\begin{center}
    \includegraphics[width=0.95\linewidth]{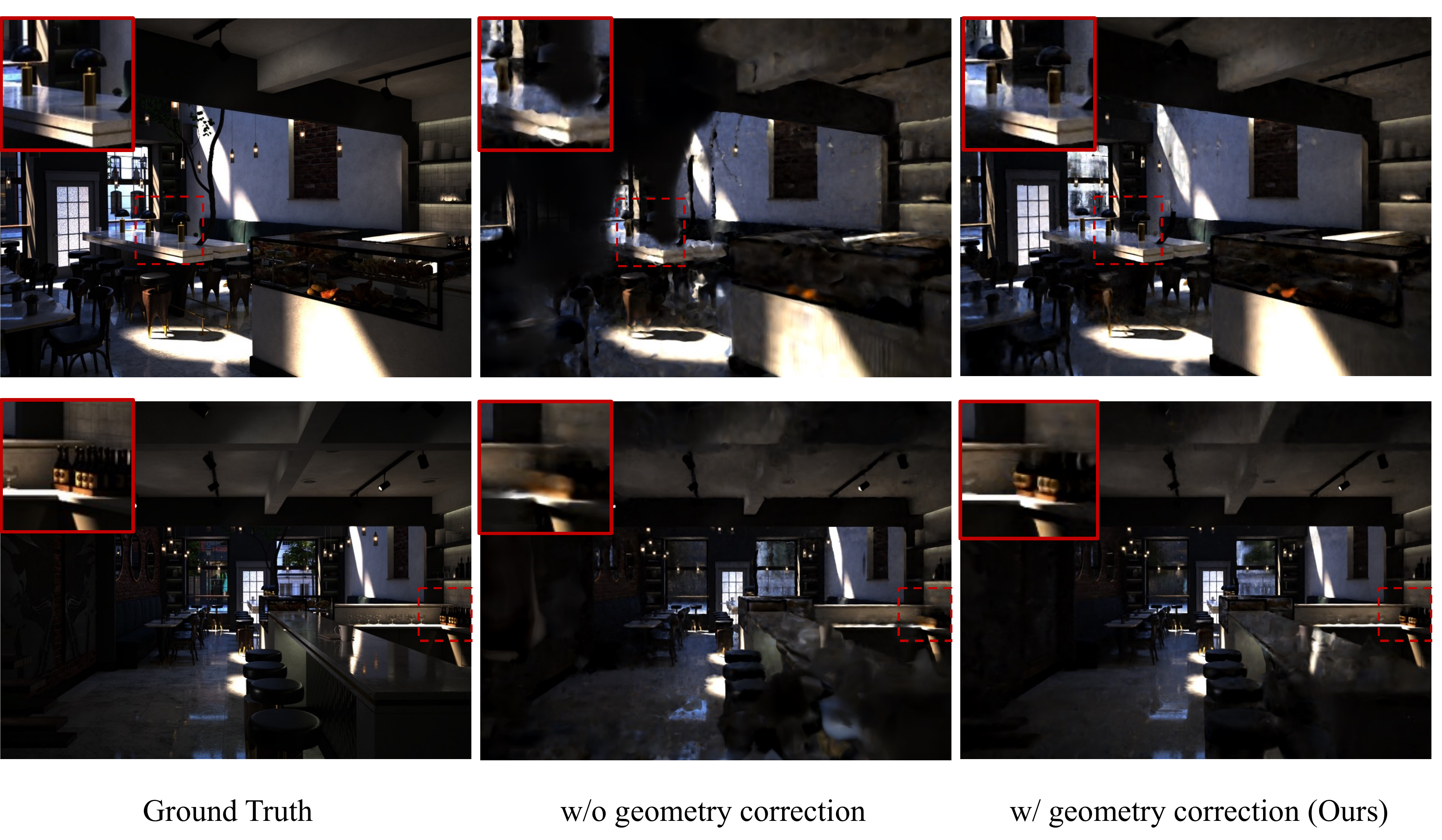}
\end{center}
\vspace{-0.1in}
\caption{\textbf{Qualitative evaluation on geometry correction.} Without geometry correction NeRF tends to generate floaters.
}
\label{fig:ab_geo}
\vspace{-0.1in}
\end{figure*}

\paragraph{Ablation  Study on our Architecture Design}
We qualitatively and quantitatively evaluate the main components of \ournerf.

Fig.~\ref{fig:ab_app} shows that without appearance decomposition, NeRF struggles to recover the glossy floor and plants.
Fig.~\ref{fig:ab_geo} shows that  without geometry correction, the NeRF needs to generate floaters in order to explain the view-dependent observations.
Quantitative evaluations are given in Tab.~\ref{table:eval2}.

\begin{table}[ht]
    
	\centering
    	\begin{tabular}{l|cccc}
    	    \multicolumn{5}{c}{ \colorbox{best1}{best} \colorbox{best2}{second-best} } \\ \hline
            Method                    & PSNR$\uparrow$          & SSIM$\uparrow$          & MAE$\downarrow$         & LPIPS$\downarrow$ \\ \hline
            NeRF         & 20.90                  & 0.84                & 0.052                  & 0.2509 \\
            Mip-NeRF 360 &21.32  & 0.82 & 0.048& 0.3399 \\
            DS-NeRF &\cellcolor{best2}26.27  & \cellcolor{best2}0.89 & \cellcolor{best1}0.03 & 0.2275 \\\hline
            w/o app. dec. &22.93  & 0.85 & 0.041 &\cellcolor{best2}0.21 \\
            w/o dec. geo. &25.88  & 0.86 & 0.0059 & 0.2299 \\\hline
        Ours      & \cellcolor{best1}27.29 & \cellcolor{best1}0.89 &  \cellcolor{best2}0.041& \cellcolor{best1}0.1975 \\\hline
        \end{tabular}
        
\rule{0pt}{0.05pt}
\caption{\textbf{Quantitative comparison and evaluation.} We compare our proposed \ournerf with representative NeRF-based methods and their variants.  }
\label{table:eval2}
\vspace{-5mm}
\end{table}

\section{Discussion and Limitations}
Our method assumes fixed lighting condition and no semi-transparent objects in a scene. In addition, we observe that, when we deal with sparse inputs and the specular highlights of a point appear in most of the inputs, such highlights may be regarded as view-independent colors, since our method does not make any assumption about the surface properties and colors. Below, we discuss some important questions related to our work:

\vspace{2mm}
\noindent\textit{Why are there floaters in sparse but not in dense inputs?} In vanilla NeRF, observation errors are backpropagated according to Eqn.~\ref{eqn:vanillanerf}, which are backpropagated equally to density and color along a given ray without any prior. With dense inputs, the strong geometry constraint from other view points can correct the density errors along a ray, and thus the view dependent observations will be correctly backpropagated to the color component. In contrast, when the number of inputs is limited, the network cannot resolve the ambiguity that the view dependent observations are caused by change of colors, or by the semi-transparent occluders, i.e., floaters. Since errors are backpropagated equally to both density and color along a ray, generating \textit{floaters} 
is more preferable by Eqn.~\ref{eqn:vanillanerf}.

\vspace{2mm}
\noindent\textit{What are the benefits of vi- and vd- color decomposition?}
Such decomposition can stabilize the solution by reducing the ambiguity in handling view-dependent observations as residual errors in ${\bf c}_\text{vd}$, while keeping the ${\bf c}_\text{vi}$ stable across multiple views, thus  leading to a reconstruction of higher quality. Additionally, in downstream tasks such as NeRF object detection~\cite{hu2022nerf} and segmentation~\cite{ren2022neural}, one may have to estimate the color of voxel features that is independent of viewpoints. Our ${\bf c}_\text{vi}$ can provide such voxel feature extraction for free without additional computations.

\vspace{2mm}
\noindent\textit{Why is the decomposition in Eqn.~\ref{eqn:blend} correct?} 
Eqn.~\ref{eqn:blend} can be considered as a simplified BRDF model, e.g., a simplified Phong model with diffuse and specular components but without normal and light. Although not entirely physically correct, this formulation can handle most view-dependent observations in the real world without resorting to estimating surface normals and incoming lighting conditions, thus providing a fast and easy way to optimize. According to our experiments, this formulation is generally applicable, and the resulting decomposition is reasonably accurate.



\section{Conclusion}
This paper proposes {\em \ournerf} for accurate 3D reconstruction and novel view rendering in complex scenes. 
By 1)  detecting and discarding view-dependent appearances in  training, followed by 2) a geometric correction procedure performed on each traced ray during inference,
\ournerf achieves unprecedented clarity and accuracy, capable of recovering intricate object details while rejecting ``floaters" when only sparse observations are given. Codes will be released.


{\small
\bibliographystyle{ieee_fullname}
\def\CVPR{IEEE/CVF Conference on Computer Vision and Pattern Recognition
  (CVPR)}\def\ECCV{ European Conference on Computer Vision
  (ECCV)}\def\ICCV{IEEE/CVF International Conference on Computer Vision
  (ICCV)}\def\NIPS{Advances in Neural Information Processing Systems
  (NeurIPS)}\def\ICML{International Conference on Machine Learning
  (ICML)}\def\ICLR{International Conference on Learning Representations
  (ICLR)}\def\WACV{IEEE/CVF Winter Conference on Applications of Computer
  Vision (WACV)}\def\CVPRW{IEEE/CVF Conference on Computer Vision and Pattern
  Recognition (CVPR) Workshops}\def\ICCVW{IEEE/CVF International Conference on
  Computer Vision (ICCV) Workshops}\def\ICRA{IEEE International Conference on
  Robotics and Automation (ICRA)}\def\TOG{ACM Transactions on Graphics
  (TOG)}\def\PAMI{IEEE Transactions on Pattern Analysis and Machine
  Intelligence (PAMI)}\def\TIP{IEEE Transactions on Image Processing
  (TIP)}\def\IJCV{International Journal of Computer Vision
  (IJCV)}\def\SIGGRAPH{ACM Transactions on Graphics
  (SIGGRAPH)}\def\SIGGRAPHASIA{ACM Transactions on Graphics (SIGGRAPH
  Asia)}\def\TOG{ACM Transactions on Graphics (TOG)}\def\threedv{International
  Conference on 3D Vision (3DV)}

}

\end{document}